\documentclass[10pt,twocolumn,letterpaper]{article}

\usepackage[pagenumbers]{cvpr} 

\definecolor{cvprblue}{rgb}{0.21,0.49,0.74}
\usepackage[pagebackref,breaklinks,colorlinks,allcolors=cvprblue]{hyperref}
\usepackage{algorithm}
\usepackage{algorithmic}
\usepackage{multirow}
\usepackage{colortbl}
\usepackage{tabularx}
\usepackage{color}
\usepackage[accsupp]{axessibility}  
\usepackage{pifont}
\usepackage{makecell}
\usepackage{multicol}
\usepackage{lipsum}
\usepackage{threeparttable}
\newcommand{\cmark}{\ding{51}\xspace}%
\newcommand{\xmarkg}{\textcolor{lightgray}{\ding{55}}\xspace}%
\def\paperID{3537} 
\def\confName{CVPR}
\def\confYear{2025}

\title{Show Me What and Where has Changed? \\ Question Answering and Grounding for Remote Sensing Change Detection}

\author{
Ke Li$^{1}$, Fuyu Dong$^{1}$, Di Wang$^{1}$\footnotemark[1], Shaofeng Li$^{1}$\footnotemark[1], Quan Wang$^{1}$, Xinbo Gao$^{1,2}$, Tat-Seng Chua$^{3}$ \\
$^{1}$ Xidian University, $^{2}$ Chongqing University of Posts and Telecommunications, \\
$^{3}$ National University of Singapore \\
\href{https://like413.github.io/CDQAG/}{https://like413.github.io/CDQAG/}
}

\begin{document}
\twocolumn[{
\renewcommand\twocolumn[1][]{#1}%
\maketitle 
\vspace{-12mm}
\begin{center} 
    \centering 
    \includegraphics[width=\textwidth]{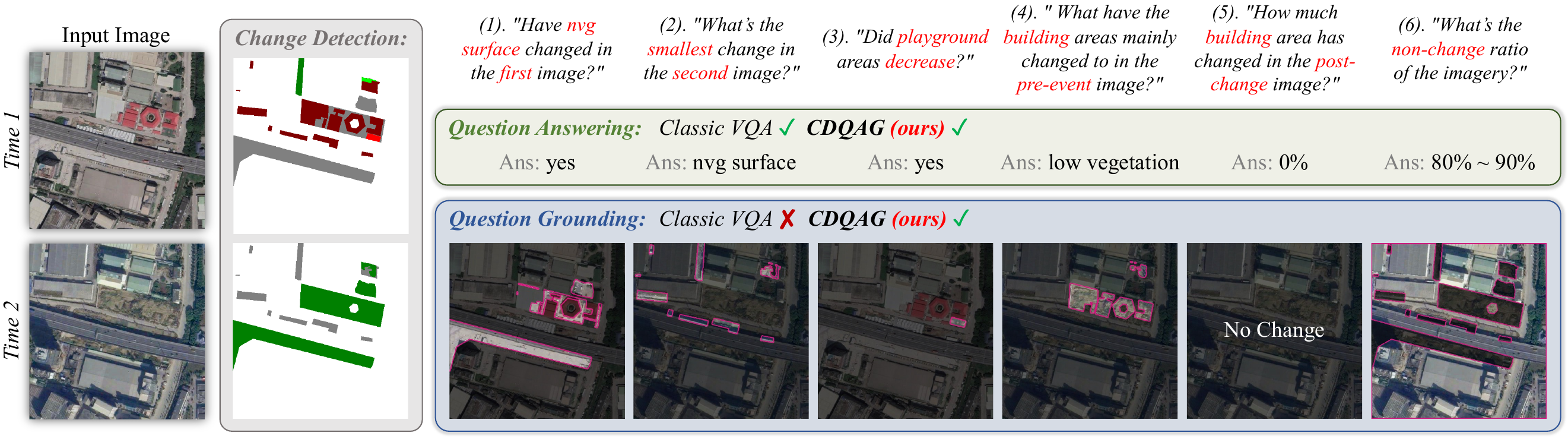} 
    \vspace{-6mm}
    \captionof{figure}{Change detection (CD) identifies surface changes from multi-temporal images.
    Classic visual question answering (VQA) only supports textual answers. In comparison, the proposed \textbf{change detection qusetion answering and grounding (CDQAG)} supports well-founded answers, \ie, textual answers (“\textit{what has changed}”) and relevant visual feedback (“\textit{where has changed}”).}
    \label{figure-1} 
\end{center}
}]

\renewcommand{\thefootnote}{\fnsymbol{footnote}}
\footnotetext[1]{Corresponding author.}

\begin{abstract}
    Remote sensing change detection aims to perceive changes occurring on the Earth's surface from remote sensing data in different periods, and feed these changes back to humans. However, most existing methods only focus on detecting change regions, lacking the capability to interact with users to identify changes that the users expect. In this paper, we introduce a new task named \textbf{C}hange \textbf{D}etection \textbf{Q}uestion \textbf{A}nswering and \textbf{G}rounding (\textbf{CDQAG}), which extends the traditional change detection task by providing interpretable textual answers and intuitive visual evidence. To this end, we construct the first CDQAG benchmark dataset, termed \textbf{QAG-360K}, comprising over 360K triplets of questions, textual answers, and corresponding high-quality visual masks. It encompasses 10 essential land-cover categories and 8 comprehensive question types, which provides a valuable and diverse dataset for remote sensing applications. Furthermore, we present \textbf{VisTA}, a simple yet effective baseline method that unifies the tasks of question answering and grounding by delivering both visual and textual answers. Our method achieves state-of-the-art results on both the classic change detection-based visual question answering (CDVQA) and the proposed CDQAG datasets. Extensive qualitative and quantitative experimental results provide useful insights for developing better CDQAG models, and we hope that our work can inspire further research in this important yet underexplored research field. The proposed benchmark dataset and method are available at \href{https://github.com/like413/VisTA}{https://github.com/like413/VisTA}.
\end{abstract}    
\vspace{-6mm} 
\section{Introduction}
\label{sec:intro}
Multi-temporal remote sensing imagery plays an essential role in the rapid acquisition of Earth's surface change information, with applications across various fields such as environmental protection, resource management, and damage assessment \cite{zheng2023scalable,seo2023self,wang2024kernel}. However, existing remote sensing change detection systems primarily focus on full-picture analysis of coverage variations, such as binary segmentation and caption generation. Such systems lack the flexibility to accommodate human instructions \cite{zheng2021change,bernhard2023mapformer}, thereby constraining the advancement of user-friendly and effective remote sensing intelligent interpretation tasks.

Prior to this paper, several studies have attempted to enhance human-computer interaction in change detection tasks through natural language question answering (\eg, CDVQA \cite{yuan2022change}). However, these methods neglect the intuitive visual explanations, \ie, the change masks referred to by the complex questions, which fail to establish a link between textual answers and the corresponding visual change regions. To this end, we propose a novel task termed \textbf{C}hange \textbf{D}etection \textbf{Q}uestion \textbf{A}nswering and \textbf{G}rounding (\textbf{CDQAG}). As illustrated in \cref{figure-1}, this task supports the simultaneous generation of textual answers and their associated pixel-level visual masks, providing an intuitive means for users to verify the answers and increasing their confidence in the reliability of the results. Therefore, exploring how to transcend the limitations of proofless answers and thoroughly investigate the semantic associations between the textual answers and visual groundings is of critical importance and constitutes the core research focus of this paper.

To propel the research of CDQAG, we construct the first benchmark dataset named QAG-360K, by extensively surveying, collecting, and standardizing 3 existing change detection datasets, generating 6,810 image pairs and over 360K high-quality \{\textit{question, answer, mask}\} triples through our automated data generation engine. Notably, the change queries include not only straightforward references (\eg, “playground”), but also more complicated description involving complex reasoning (\eg, “the land cover category with the smallest change” or “the area of land cover that increased”). \cref{figure-1} shows some examples of CDQAG task. For example, given “How much building area has changed in the post-change image?”, the output should provide both the proportion range of “building” and the corresponding change mask. Another example is “What is the smallest change in the second image?”, where the answer necessitates perceiving and evaluating changes across different categories to generate a correct textual-visual answer. \cref{secdata} provides a comprehensive description and statistical analysis of the proposed dataset.

\begin{table}[!t]
\centering
\resizebox{\linewidth}{!}{
    \begin{tabular}{lcccc}
    \toprule
    Datasets                          & Images          & Questions          & Textual-Ans.     & Visual-Ans.  \\       
    \midrule
    LEVIR-CD \cite{chen2020spatial}   & 637             & {\xmarkg}          & {\xmarkg}        & {\xmarkg}                   \\
    SECOND \cite{yang2021asymmetric}  & 4,662           & {\xmarkg}          & {\xmarkg}        & {\xmarkg}                   \\
    CDVQA \cite{yuan2022change}       & 2,968           & 122K               & {\cmark}         & {\xmarkg}                   \\
    \textbf{QAG-360K}                 & 6,810           & 360K               & {\cmark}         & {\cmark}                    \\
    \bottomrule
    \end{tabular}}
    \vspace{-3mm}
    \caption{Comparison among different remote sensing change detection related datasets, including common segmentation datasets \cite{chen2020spatial,yang2021asymmetric}, visual question answering dataset \cite{yuan2022change}, and the proposed \textbf{QAG-360K}.}
    \label{table-1}
    \vspace{-5mm}
\end{table}

To accomplish this task, we develop a powerful benchmark framework called VisTA, which aims to tackle two key challenges: (1) How to reason about complex questions jointly with change images? 
(2) How to uncover the intrinsic relationships between the textual answers and the corresponding visual answers? 
To solve the first challenge, the Multi-Stage Reasoning Module is designed to adaptively enhance cross-modal information interaction and fusion by jointly leveraging pixel-level visual representations and fine-grained textual features. Notably, based on the observation that answer grounding typically depend on specific semantic cues in the question or answer, we introduce a question-answer selection module to further refine the reasoning process. While the Text-Visual Answer Decoder is proposed to address the second challenge, which employs a text-to-pixel contrastive learning strategy to achieve fine-grained alignment between textual answers and visual masks.

In summary, our contributions are as follows:
\begin{itemize}[left=1pc]
	\item{
        We propose a novel CDQAG task for remote sensing change detection. Unlike classic VQA (see \cref{figure-1}), CDQAG not only generates textual answers but also provides pixel-level visual evidence, which is crucial for developing reliable remote sensing change detection systems.
        }
	\item{
        We construct the first CDQAG benchmark dataset QAG-360K, containing over 360K \{\textit{question, answer, mask}\} triples that are not only diverse in land-cover categories but contain comprehensive questions. This benchmark is essential for evaluation and encourages the community to further explore the CDQAG task.
        }
	\item{
        We present a solid CDQAG baseline method VisTA, which surpasses the latest visual question answering and visual grounding methods, achieving state-of-the-art performance on both the classic CDVQA and the porposed QAG-360K datasets.
        }
\end{itemize}
\begin{figure*}[t]
  \centering
  \hfill
    \begin{subfigure}[t]{0.20\linewidth}
      \centering
      \includegraphics[width=\textwidth]{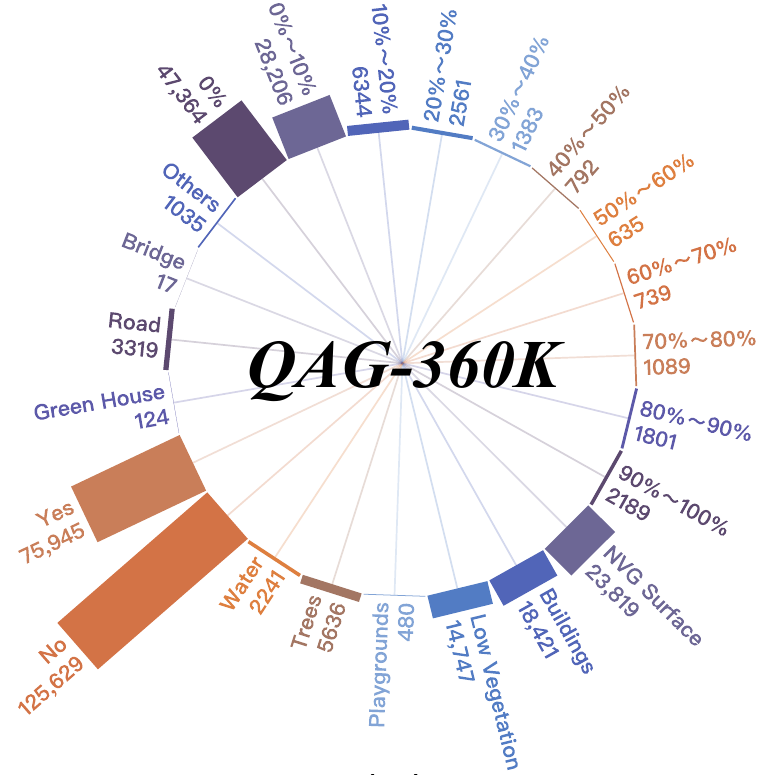}
      \vspace{-6mm}
      \caption{}
      \label{fig:statisticsdata1}
    \end{subfigure}
  \hfill
    \begin{subfigure}[t]{0.3\linewidth}
      \centering
      \includegraphics[width=\textwidth]{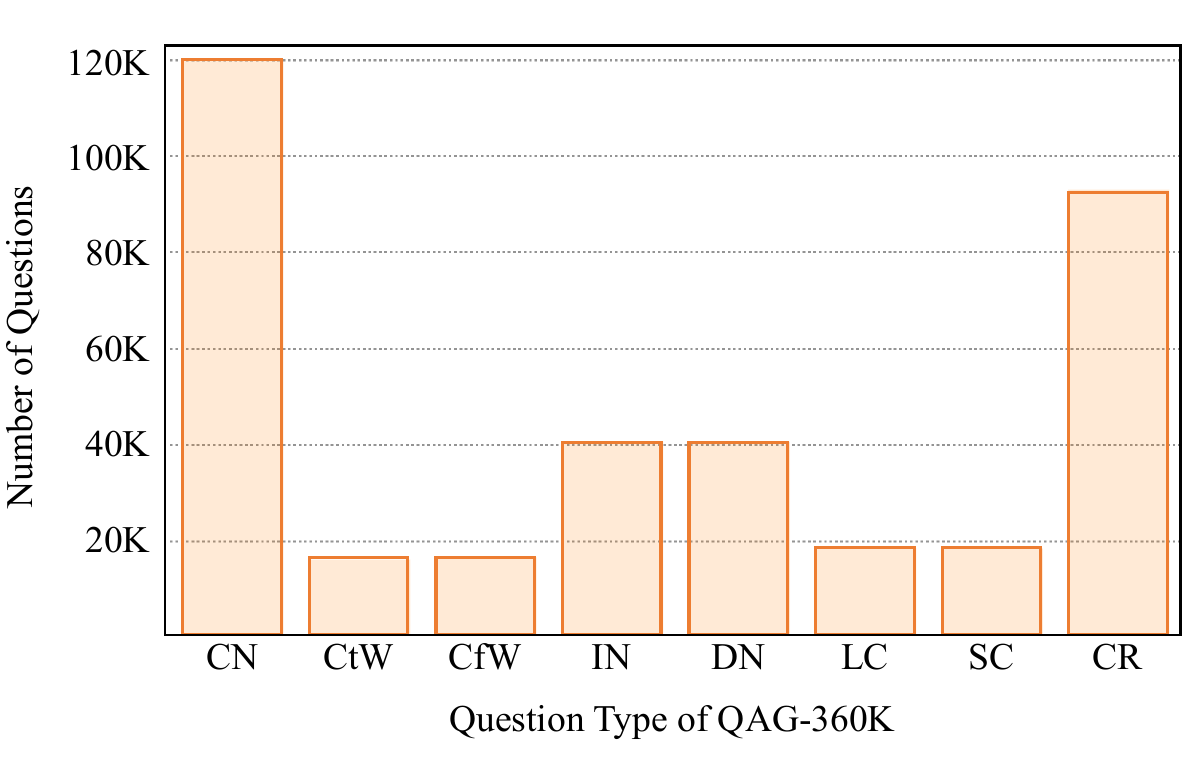}
      \vspace{-6mm}
      \caption{}
      \label{fig:statisticsdata2}
    \end{subfigure}
  \hfill
    \begin{subfigure}[t]{0.18\linewidth}
      \centering
      \includegraphics[width=\textwidth]{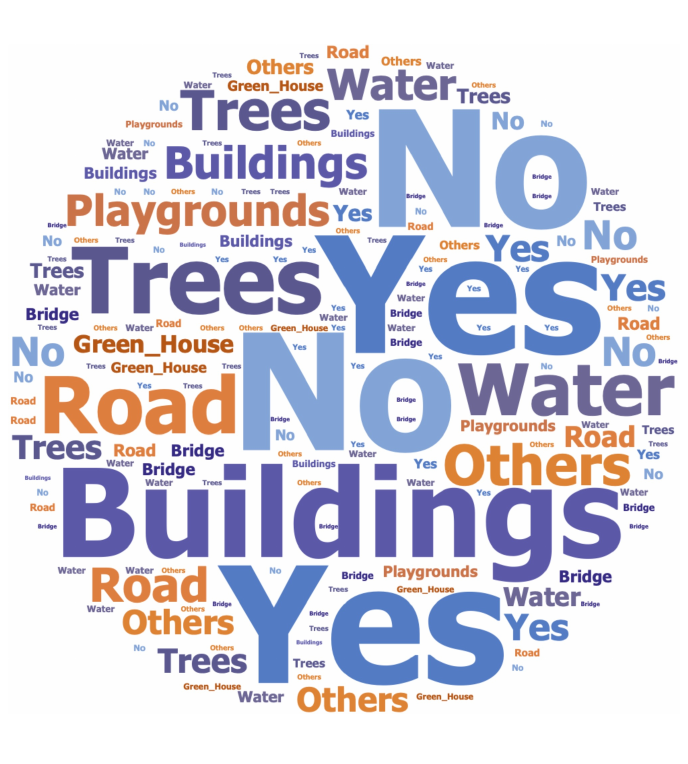}
      \vspace{-6mm}
      \caption{}
      \label{fig:statisticsdata3}
    \end{subfigure}
  \hfill
    \begin{subfigure}[t]{0.3\linewidth}
      \centering
      \includegraphics[width=\textwidth]{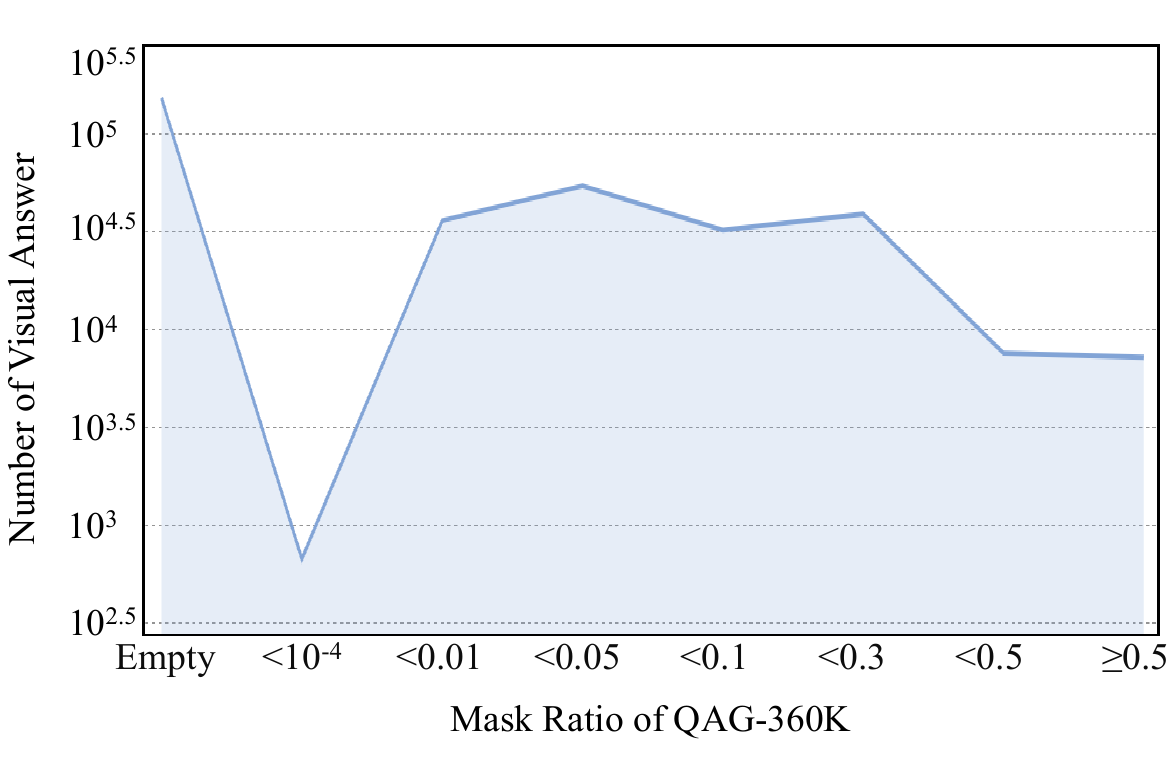}
      \vspace{-6mm}
      \caption{}
      \label{fig:statisticsdata4}
    \end{subfigure}
  \hfill
  \vspace{-3mm}
  \caption{The proposed QAG-360K benchmark statistics: (a) the frequency of answer categories; (b) the distribution of question types (CN: change or not; CtW: change to what; CfW: change from what; IN: increase or not; DN: decrease or not; LC: largest change; SC: smallest change; CR: change ratio); (c) the word cloud of all questions; (d) the distribution of area ratio for visual masks.}
  \label{fig:statistics}
  \vspace{-5mm}
\end{figure*}

\section{Related work}
\label{sec2}

\noindent \textbf{Remote Sensing Change Detection.} 
Remote sensing change detection involves identifying changes in land-cover information and plays a significant role in urban planning and environmental monitoring. By the type of change, it can be categorized into binary change detection \cite{dong2024changeclip, wu2023fully, bernhard2023mapformer} and semantic change detection \cite{ding2024joint, zheng2022changemask,toker2022dynamicearthnet}. The former only determines whether change have occurred between the two time periods, while the latter can further provide more detailed information about the categories of land-cover changes. Thus, we collecte existing semantic change detection datasets \cite{tian2020hi, yang2021asymmetric, chen2020spatial}, to prepare for CDQAG task.

\noindent \textbf{Visual Question Answering.} 
As a pivotal task in multimodal research, visual question answering (VQA) requires models to respond to natural language queries based on images, making it a focal point in the domain of vision-language research. In recent years, the release of numerous benchmark datasets, including VQA \cite{goyal2017making}, COCO-QA \cite{ren2015exploring}, Visual Genome \cite{krishna2017visual}, and CLEVR \cite{johnson2017clevr}, has significantly accelerated progress in this field. Simultaneously, VQA also demonstrates its enormous potential in many applications such as education, healthcare, and remote sensing \cite{ma2024robust,wang2024earthvqa,Hu_2024_CVPR}. In particular, the introduction of CDVQA \cite{yuan2022change} has brought VQA into the domain of change detection, establishing a baseline dataset and method that expand the research frontiers of VQA while providing fresh perspectives for multimodal analysis in remote sensing.

\noindent \textbf{Visual Grounding.} 
Visual grounding (VG) focuses on localizing specific objects within images based on natural language descriptions, and can be typically divided into two subtasks: referring expression comprehension (REC) \cite{ye2022shifting,deng2023transvg++,yang2022improving} and referring expression segmentation (RES) \cite{xu2023bridging,yang2022lavt}. REC aims to predict the bounding boxes of target objects from linguistic descriptions, while RES requires the generation of  pixel-level segmentation masks of the referred targets. Recently, VG in remote sensing \cite{li2024language, liu2024rotated} has gained increasing attention for its ability to intuitively highlight targets of interest against irrelevant backgrounds.

\noindent \textbf{Question Answering and Grounding.}
As a hybrid task that combines VQA and VG, question answering and grounding (QAG) requires vision-language models to supply visual evidence while answering questions \cite{zhu2016visual7w,gurari2018vizwiz}. Unlike traditional \textit{post-hoc} interpretability methods \cite{ribeiro2016should,selvaraju2017grad}, QAG addresses concerns about “\textit{whether answer reasoning is based on correct visual evidence}" by providing intuitive explanations, making it crucial for practical applications that require high safety and reliability. To support this capability, multiple VQA datasets now provide grounding labels such as VQS \cite{gan2017vqs}, GQA \cite{hudson2019gqa}, and TVQA+ \cite{lei2019tvqa+}. However, QAG remains largely unexplored in the field of remote sensing. To bridge this gap, we introduce CDQAG, the first benchmark dataset for the remote sensing change detection task, aiming to respond accurately and reliably to user inquiries about surface changes in different periods. 
The dataset comprises over 6.8K image pairs and 360K questions across 10 different land-cover categories.
Compared with QAG tasks in natural scenes, remote sensing change detection involves more complex geospatial data and implicit reasoning relationships, imposing higher demands on the perception and interpretability capabilities of the models. Furthermore, existing visual responses are usually provided in the form of object-level bounding boxes \cite{zhu2016visual7w,urooj2021found}. To meet the distinct requirements of change detection tasks, we provide pixel-level visual answers, enabling more precise and fine-grained analysis of changes.

\section{Dataset Construction}
\label{secdata}
We now introduce our task and dataset, which provide well-founded answers for land-cover change-related questions. We first define the task of change detection visual question answering and grounding (CDQAG) and then present the first benchmark dataset QAG-360K.

\subsection{Task Definition}
The input of the CDQAG task contains a pair of remote sensing images $T_1$ and $T_2$, captured in the same location but at a different time, along with a question $Q$. The output is a textual answer $A$ and a corresponding visual segmentation $S$. Unlike classic VQA methods that provide only natural language responses, CDQAG can offer both textual answers and correlative visual explanations (as shown in \cref{figure-1}), which is critical for reasonable remote sensing interpretation.

\begin{figure*}[!htp]
	\centerline{\includegraphics[width=\linewidth]{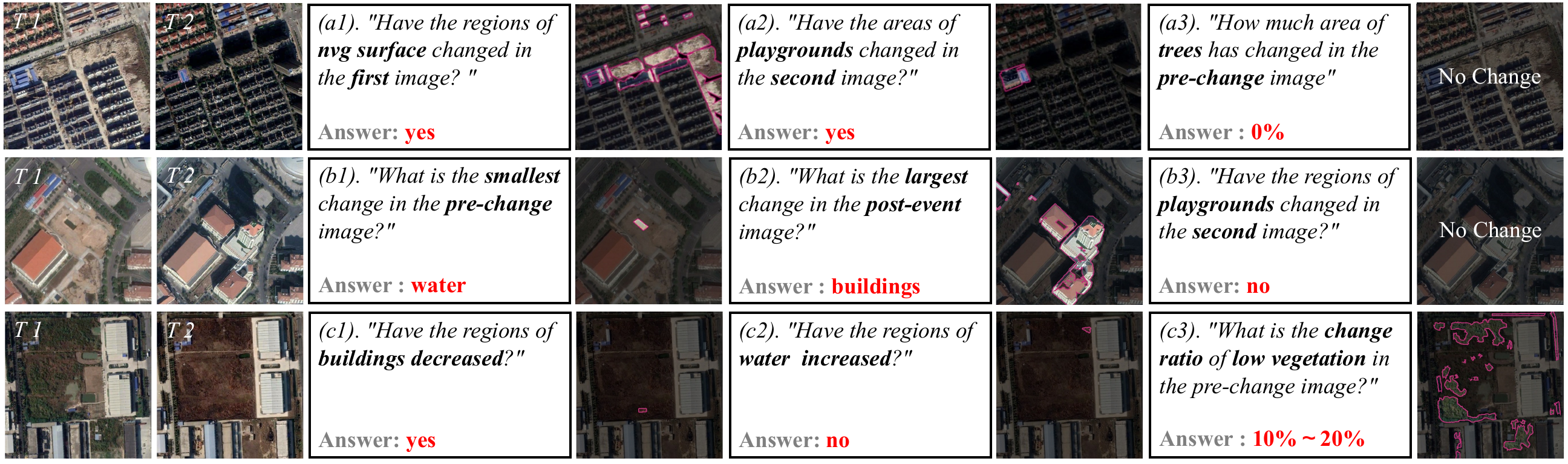}}
  \vspace{-3mm}
	\caption{Examples of the proposed QAG-360K dataset.}
	\label{examples}
  \vspace{-5mm}
\end{figure*}

\subsection{QAG-360K: A Large-scale CDQAG Dataset}
To address the gap in this critical yet underexplored area, we introduce QAG-360K, the first benchmark dataset tailored for the CDQAG task. We collected a high-quality set of remote sensing images from existing binary and semantic change detection datasets, including Hi-UCD \cite{tian2020hi}, SECOND \cite{yang2021asymmetric}, and LEVIR-CD \cite{chen2020spatial}. 
These data cover 24 different regions across cities in Estonia, China and the United States, with spatial resolution ranging from 0.1 to 3.0 meters, providing diverse geographical scenarios.
We filtered 6,810 remote sensing image pairs from these datasets as part of QAG-360K, including 10 land-cover categories, all of which are equipped with semantic masks. Additionally, we developed a specialized Triplet Generation Engine (in Appendix A) to automatically produce change-related questions, textual answers, and corresponding visual masks.

\noindent \textbf{Question Raised.} 
To encompass a broad range of change scenarios, we expanded the change-related questions into \textit{\textbf{8C Questions}}, covering the critical change detection types: \textit{change or not? change to what? change from what? increase or not? decrease or not? largest change? smallest change? change ratio?} Moreover, to ensure that the phrasing of the questions conforms to natural communication, we utilized the large language model (LLM) to generate an average of 20 question templates for each change type and then manually selected the most reasonable 5 for querying. The final questions average 9.5 words in length, with lengths ranging from 4 to 15 words. Each pair of remote sensing images contains an average of 53 triples. \cref{fig:statisticsdata2} and \cref{fig:statisticsdata3} illustrate the distribution of all question types and the word clouds of land-cover categories, respectively. Detailed definitions and templates for each question type are provided in the Appendix A.2.

\noindent \textbf{Answers Generation.}
For each question type in the \textit{\textbf{8C Questions}}, the answer generation follows different judgment rules. Specifically, existing change detection datasets offer mask annotations, which allow us to automatically generate textual answers and the corresponding pixel-level groundings based on predefined rules. The answer generation rules and specific procedures for each question type are detailed in the Appendix A.2. \cref{fig:statisticsdata4} illustrates the distribution of mask area ratios for each visual answer. CDQAG introduces four critical and distinct attributes that are essential for practical applications:
\begin{itemize}
  \item[1)]{
    \textbf{Temporal Semantic Correlation.}
    CDQAG involves comparing and analyzing multi-temporal remote sensing images. For instance, \cref{examples}(a1) and \cref{examples}(a2) pose two questions regarding pre-change and post-change conditions, respectively. The model is required to accurately discern and associate temporal semantic features from multi-temporal imagery, as dictated by natural language queries, in order to correctly evaluate and interpret changes in the target region.
  }
  \item[2)]{
    \textbf{Complex and Implicit Reasoning.}
    In contrast to visual grounding tasks with clear-cut references, CDQAG typically involves complex reasoning about implicit information. For example, in \cref{examples}(b1) and \cref{examples}(b2), adjectives such as “largest” and “smallest” are used, requiring the model to match these descriptions to the corresponding areas. Furthermore, the QAG-360K dataset incorporates diverse ordinal expressions (\eg, “before,” “pre-change,” and “first”) to indicate temporal changes, which necessitates that the model comprehend such terms to execute logical reasoning effectively.
  }
  \item[3)]{
    \textbf{Fragmented Spatial Masks.}
    Unlike natural images, which generally exhibit object-level semantic features, certain land-cover categories in remote sensing imagery are spread across multiple regions, leading to fragmented and discontinuous segmentation masks. \cref{examples}(c3) illustrates an example of proportional change in low vegetation. Moreover, when changes are extremely subtle, it poses a greater challenge for the model to accurately identify and segment them.
  }
  \item[4)]{
    \textbf{Inconspicuous Visual Cues.}
    In certain instances, the query target may remain unchanged or may not appear at all, as exemplified by \cref{examples}(a3) and \cref{examples}(b2). Under such cases, the model must generate an empty mask to accurately indicate that no actual change has taken place, rather than resorting to guesswork. This imposes higher demands on the model's reasoning capabilities, contextual comprehension, and capacity to handle uncertainty.
  }
\end{itemize}
\begin{figure*}[!t]
	\centerline{\includegraphics[width=\linewidth]{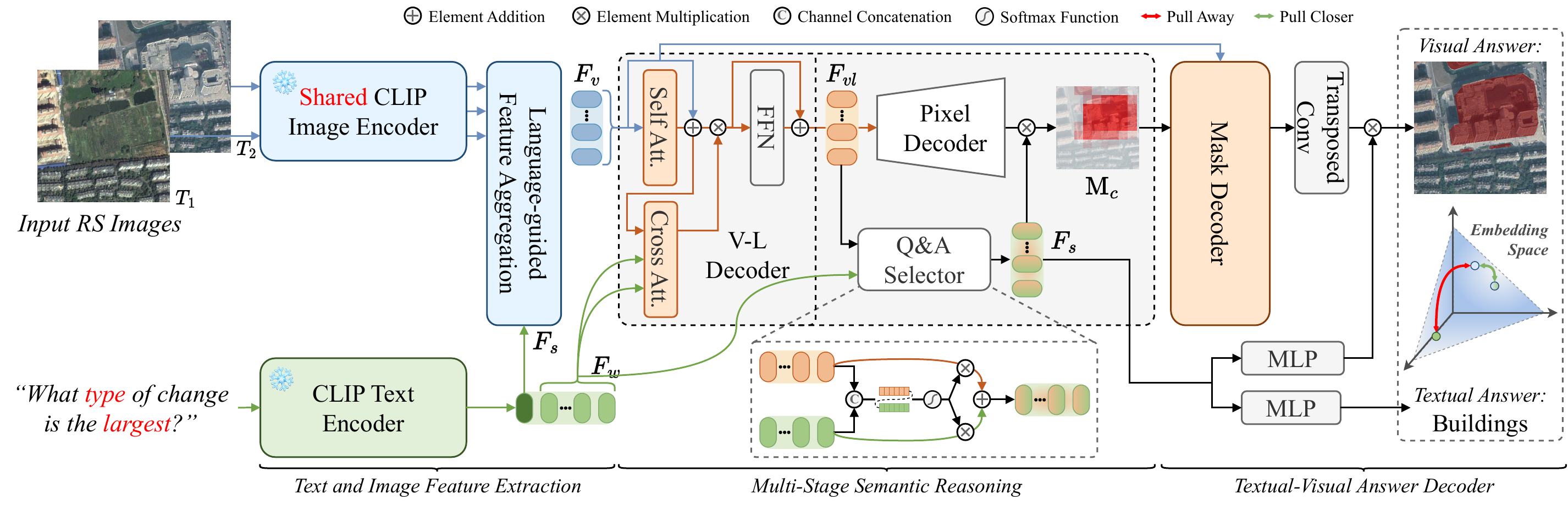}}
  \vspace{-3mm}
	\caption{The architecture of our VisTA model as a simple baseline for the CDQAG task.
  Firstly, the given two remote sensing images and a question are encoded into vision features $F_v$ and language features $[F_s,F_w]$, respectively. $F_v$ and $F_w$ are fed into a vision-language decoder to produce the refined multimodal features $F_{vl}$. 
  Next, the Q\&A selector is used to generate Q\&A features $F_s$. Subsequently, $F_s$ is activated as a selection weight to filter the pixel decoder's output, resulting in coarse mask $\text{M}_c$. Finally, the textual-visual answer decoder is employed to predict the textual answer and the corresponding visual answer.}
	\label{architecture}
  \vspace{-5mm}
\end{figure*}
These characteristics are pervasive throughout our dataset, ensuring that it maintains a balanced level of complexity. Furthermore, more examples from the QAG-360K dataset are provided in the Appendix B.2.
\section{Method}
\label{sec3}
Next, we provide a detailed description of our proposed CDQAG model VisTA, which delivers reliable textual answers and precise visual feedback. As shown in \cref{architecture}, our framework consists of three primary components, which are explained below.

\subsection{Text and Image Feature Extraction}
\textbf{Text Encoder.}
Given a question $Q \in \mathbb{R}^L$, we leverage a CLIP pre-trained Transformer \cite{radford2021learning} to extract text features $F_t \in \mathbb{R}^{L \times C}$, where $L$ denotes the question length and $C$ represents the number of feature channels. The input text sequence is surrounded by two special tokens [SOS] and [EOS], where [EOS] is subsequently activated as a sentence-level representation $F_s \in \mathbb{R}^C$.

\noindent \textbf{Image Encoder.}
For the two remote sensing images $T_1 \in \mathbb{R}^{H \times W \times 3}$ and $T_2 \in \mathbb{R}^{H \times W \times 3}$, we extract multi-scale visual features $F_{v_i}^{1} \in \mathbb{R}^{H_i \times W_i \times C_i}$ and $F_{v_i}^{2} \in \mathbb{R}^{H_i \times W_i \times C_i}$ using two ResNets \cite{he2016deep} with shared weights. Here, $i \in \{3,4,5\}$ denotes the $i$-th stage of visual backbone, and $H_i = H/2^i$ and $W_i = W/2^i$ are the corresponding resolutions. The original image dimensions are $H$ and $W$, respectively. The ResNet is also pre-trained on CLIP, enhancing its visual representation capabilities. To obtain multi-scale change features $F_{c_i} \in \mathbb{R}^{H_i \times W_i \times C_i}$, we concatenate the visual features from both ResNets and utilize a $1 \times 1$ convolutional layer to adjust the channel dimensions:
\begin{equation}
    \label{tab:equ1}
    F_{c_i} = \text{Conv}([F_{v_i}^1, F_{v_i}^2]),
\end{equation}
where $[ , ]$ indicates concatenation.

\noindent \textbf{Language-Guided Feature Aggregation.}
To efficiently fuse textual and change-related visual features, we design a module that initiates cross-modal feature fusion:
\begin{equation}
    \label{tab:equ2}
    F_{m_5} = \text{Conv}(F_{c_5}) \cdot \text{Linear}(F_s),
\end{equation}
where $F_{m_5}$ represents the fused multi-modal features, and $\text{Linear}(\cdot)$ refers to a multi-layer perceptron that adjusts the dimensionality of the text features. 
We then feed $F_{m_5}$, $F_{c_4}$, and $F_{c_3}$ into a common Feature Pyramid Network \cite{lin2017feature}, denoted as $\mathcal{F} _{\text{FPN}}(\cdot)$, to extract multi-scale language-guided features:
\begin{equation}
    \label{tab:equ3}
    F_{m_5}, F_{m_4}, F_{m_3} = \mathcal{F} _{\text{FPN}}(F_{m_5}, F_{c_4}, F_{c_3}).
\end{equation}
Subsequently, we aggregate the three multi-modal features via a convolutional layer:
\begin{equation}
    \label{tab:equ4}
    F_m = \text{Conv}([\text{Conv}(F_{m_5}), F_{m_4}, \text{DeConv}(F_{m_3})]),
\end{equation}
where $\text{DeConv}(\cdot)$ signifies the deconvolutional layer. For subsequent cross-modal interaction, we flatten the output $F_m \in \mathbb{R}^{\frac{H}{16} \times \frac{W}{16} \times C}$ to obtain the preliminary visual features $F_v \in \mathbb{R}^{N \times C}$, where $N = H/16 \times W/16$.

\subsection{Multi-Stage Semantic Reasoning}
To address the complex reasoning demands of the CDQAG task, we propose a multi-stage reasoning module that facilitates fine-grained cross-modal information interaction. As shown in \cref{architecture}, given the pixel-level visual features $F_v$ and word-level text features $F_w$, we construct a vision-language decoder to enable dense cross-modal interaction, producing a series of refined multi-modal features $F_{vl} \in \mathbb{R}^{N \times C}$. The interaction can be formalized as:
\begin{equation}
    \label{tab:equ5}
    \begin{split}
        & F_v' = F_v + \text{SA}(F_v), \\
        & F_{vl}' = \text{CA}(F_v', F_w) \cdot F_v', \\
        & F_{vl} = \text{FFN}(F_{vl}' \cdot F_v') + F_{vl}', \\
    \end{split}
\end{equation}
where $\text{SA}(\cdot)$ and $\text{CA}(\cdot)$ denote the multi-head self-attention and cross-attention layers, and $\text{FFN}(\cdot)$ refers to the feed-forward network. Since some questions or answers may lack explicit references, we introduce a Question and Answer Selector, allowing the model to dynamically select question-related or answer-related text representations. Rather than directly adding or concatenating these two features, we adaptively merge the multi-modal features $F_{vl}$ and text features $F_w$ via a soft attention mechanism, generating selection features $F_s$. This process can be expressed as:
\begin{equation}
    \label{tab:equ6}
    \begin{split}
        \alpha = \frac{e^{F_{vl}}}{e^{F_{vl}} + e^{F_w}},& \quad \beta = 1 - \alpha, \\
        F_s = \alpha F_{vl} & + \beta F_w.
    \end{split}
\end{equation}
We then activate $F_s$ as a selection weight to filter out irrelevant change regions, which is formulated as:
\begin{equation}
    \label{tab:equ7}
    \text{M}_{c} = \sigma(F_s)\otimes \mathcal{F} _{\text{PD}}(F_{vl}),
\end{equation}
where $\sigma(\cdot)$ represents the sigmoid function, and $\mathcal{F}_{\text{PD}}(\cdot)$ is a pixel decoder that transforms $F_{vl}$ into a coarse mask $\text{M}_{c}$.

\subsection{Text-Visual Answer Decoder}
We leverage the established visual coarse mask $\text{M}_{c}$ and the quastion-answer features $F_s$ for the ﬁnal predictions. Specifically, in the  visual branch, the coarse mask $\text{M}_{c}$ serves as a dense visual prompt and is fed into the Mask Decoder $\mathcal{F} _{\text{MD}}(\cdot)$, along with original visual features $F_v$, to generate a more precise answer grounding:
\begin{equation}
    \label{tab:equ8}
    \tilde{F_\text{M}} = \mathcal{F} _{\text{MD}}(\text{M}_{c}, F_v).
\end{equation}
The Mask Decoder, comprising two consecutive Two-Way Attention Blocks \cite{kirillov2023segment}, is designed to establish the pixel-level mapping between $\text{M}_{c}$ and $F_v$. For textual answer, we apply a two-layer MLP followed by softmax activation to perform classification predictions. Notably, to enhance semantic consistency between visual and textual answers, we split and reshape the text features into a weight $W \in \mathbb{R}^{D \times K \times K}$ and a bias $b \in \mathbb{R}^D$, where $K$ is the kernel size of the convolutional layer. This enables it to function as the kernel and bias for a 2D convolutional layer, transforming the visual features into the final precise binary mask $\text{M}$. The implementation can be formulated as follows:
\begin{equation}
    \label{tab:equ9}
    \begin{split}
        W, b =& S\&R(\text{Linear}(F_s)), \\
        \text{M} =& W*\tilde{F_\text{M}} +b, \\
    \end{split}
\end{equation}
where $S\&R(\cdot)$ represents split and reshape, and $*$ is convolution operation.

\subsection{Training Objective}
Following prior works \cite{lobry2020rsvqa}, we employ the standard cross-entropy (CE) loss for textual answer classification, denoted as $\mathcal{L}_{txt}$. 
For visual evidence, it is noteworthy that we incorporate a text-to-pixel contrastive loss into our training objective to ensure alignment with the corresponding textual answer.
Specifically, the text-to-pixel contrastive loss measures the similarity between the textual answer $F_{ta}$ and the pixel-level visual answer $F_{va}$ using a dot product. The contrastive loss $\mathcal{L} _{con}$ is defined as:
\begin{equation}
    \begin{split}
        \mathcal{L} _{con}(F_{ta}, F_{va}) &= \frac{1}{|\mathcal{P} \cup \mathcal{N}|} \sum_{i \in \mathcal{P} \cup \mathcal{N}} \mathcal{L} _{con}^i(F_{ta}, F_{va}^i), \\
        s.t.\ \mathcal{L} _{con}^i(F_{ta}, F_{va}^i) &= 
        \begin{cases} 
            -\log \sigma(F_{ta} \cdot F_{va}^i), & i \in \mathcal{P}, \\
            -\log(1 - \sigma(F_{ta} \cdot F_{va}^i)), & i \in \mathcal{N},
        \end{cases}
    \end{split}
\end{equation}
where $\mathcal{P}$ and $\mathcal{N}$ represent the positive and negative classes, respectively, in the ground truth. $|\mathcal{P} \cup \mathcal{N}|$ is the cardinality, and $\sigma(\cdot)$ is the sigmoid function. Our final objective function is:
\begin{equation}
    \mathcal{L}  = \lambda_1 \mathcal{L}_{txt} + \lambda_2 \mathcal{L}_{con},
    \label{eq:loss}
\end{equation}
where \(\lambda_1\) and \(\lambda_2\) are hyperparameters that balance the three losses during training.

\begin{table*}[!t]
\centering
\renewcommand{\arraystretch}{0.9}
\resizebox{\linewidth}{!}{
\begin{tabular}{c|c|ccccccccc}
    \specialrule{.1em}{.05em}{.05em}
    \multicolumn{2}{c|}{\multirow{2}{*}{Method}} & CRIS \cite{wang2022cris}    & LAVT \cite{yang2022lavt}   & CGFormer \cite{tang2023contrastive}   & ETRIS \cite{xu2023bridging}  & RSVQA \cite{lobry2020rsvqa}    & CDVQA \cite{yuan2022change}   & SOBA \cite{wang2024earthvqa}      & VisTA   & $\dagger$VisTA         \\ 
    \multicolumn{2}{c|}{}                        & \textcolor{gray}{CVPR'22}   & \textcolor{gray}{CVPR'22}  & \textcolor{gray}{CVPR'23}             & \textcolor{gray}{ICCV'23}    & \textcolor{gray}{TGRS'21}      & \textcolor{gray}{TGRS'22}     & \textcolor{gray}{AAAI'24}         & \textbf{Ours}        & \textbf{Ours}         \\ 
    \specialrule{.1em}{.05em}{.05em}
    \multicolumn{2}{c|}{Backbone}                & Res-101 & Swin-B & Res-101& ViT-B/16 & Res-101& ViT-B/16 & Swin-T & Res-101& Res-101\\ 
    \midrule
    \multirow{10}{*}{\rotatebox{90}{Textual Answer}} 
                            & CN                 & 82.29 & 83.43 & 82.90 & 84.08 & 81.73 & 82.30 & 84.74 & \underline{86.37} & \textbf{87.02} \\ 
                            & CtW                & 58.49 & 59.94 & 59.44 & 60.89 & 57.39 & 58.49 & 62.12 & \underline{66.48} & \textbf{68.34} \\ 
                            & CfW                & 58.57 & 59.37 & 59.46 & 60.22 & 57.60 & 60.34 & 60.95 & \underline{66.81} & \textbf{69.92} \\ 
                            & IN                 & 79.59 & 81.66 & 81.02 & 82.39 & 75.28 & 76.02 & 81.08 & \underline{85.33} & \textbf{87.05} \\ 
                            & DN                 & 80.32 & 81.08 & 81.44 & 82.99 & 78.18 & 77.66 & 80.69 & \underline{86.35} & \textbf{87.43} \\ 
                            & LC                 & 52.16 & 52.81 & 53.52 & 55.25 & 47.89 & 49.19 & 56.91 & \underline{64.88} & \textbf{68.08} \\ 
                            & SC                 & 31.76 & 32.52 & 32.35 & 33.11 & 29.81 & 28.91 & 33.87 & \underline{40.30} & \textbf{41.57} \\ 
                            & CR                 & 66.87 & 69.00 & 68.45 & 69.50 & 63.73 & 61.81 & 69.76 & \underline{72.76} & \textbf{74.79} \\ 
                            \cmidrule(r){2-11}
                            & \textbf{AA}        & 63.76 & 64.98 & 64.82 & 66.06 & 61.45 & 62.46 & 66.27 & \underline{71.16} & \textbf{73.03} \\ 
                            & \textbf{OA}        & 69.50 & 70.89 & 70.58 & 71.78 & 67.35 & 68.60 & 71.98 & \underline{75.76} & \textbf{77.35} \\ 
                            \cmidrule(r){1-11}
    \multicolumn{2}{c|}{$\flat$ OA Decl.}     &\textcolor{blue}{-7.85} &\textcolor{blue}{-6.46} &\textcolor{blue}{-6.77} &\textcolor{blue}{-5.57} &\textcolor{blue}{-10.0} &\textcolor{blue}{-8.75} & \textcolor{blue}{-5.37}    &\textcolor{blue}{-1.59} & -    \\
    \specialrule{.1em}{.05em}{.05em}
    \multirow{10}{*}{\rotatebox{90}{Visual Answer}} 
                            & CN                 &21.70  &24.82  &23.05  &28.50  &13.84  &14.96  &26.92  &\underline{39.00}& \textbf{44.18} \\ 
                            & CtW                &23.46  &26.12  &24.02  &27.16  &10.15  &10.94  &26.66  &\underline{33.84}& \textbf{40.41} \\ 
                            & CfW                &27.68  &30.32  &27.93  &31.57  &10.76  &11.97  &31.10  &\underline{37.44}& \textbf{43.30} \\ 
                            & IN                 &28.66  &30.71  &27.62  &33.56  &17.40  &17.61  &32.68  &\underline{40.10}& \textbf{45.74} \\ 
                            & DN                 &25.97  &29.00  &26.47  &30.56  &17.70  &18.61  &29.88  &\underline{40.86}& \textbf{46.55} \\ 
                            & LC                 &36.44  &37.53  &36.85  &39.66  &17.27  &18.11  &38.88  &\underline{43.68}& \textbf{48.71} \\ 
                            & SC                 &9.89   &14.94  &10.52  &17.88  &1.39   &1.76   &16.92  &\underline{21.04}& \textbf{24.78} \\ 
                            & CR                 &54.98  &58.46  &56.17  &59.57  &46.84  &47.31  &58.86  &\underline{63.03}& \textbf{65.86} \\ 
                            \cmidrule(r){2-11}
                            & \textbf{mIoU}      &28.60  &31.49  &29.08  &33.56  &16.91  &17.66  &32.74  &\underline{39.87}& \textbf{44.94} \\ 
                            & \textbf{oIoU}      &37.42  &40.28  &38.83  &42.56  &28.54  &29.33  &41.83  &\underline{47.98}& \textbf{52.13} \\ 
    \cmidrule(r){1-11}
    \multicolumn{2}{c|}{$\flat$ oIoU Decl.}     &\textcolor{blue}{-14.71} &\textcolor{blue}{-11.85} &\textcolor{blue}{-13.30} &\textcolor{blue}{-9.57} &\textcolor{blue}{-23.59} &\textcolor{blue}{-22.80} &\textcolor{blue}{-10.30} &\textcolor{blue}{-4.15} & -    \\                 
    \specialrule{.1em}{.05em}{.05em}
\end{tabular}}
\vspace{-3mm}
\caption{Comparisons with the state-of-the-art methods on the proposed \textbf{QAG-360K} test set. ``$\flat$ Decl.'' indicates performance decline. The best and second best performance are highlighted in \textbf{bold} and \underline{underline}. `$\dagger$' represents using two weight-shared visual backbone.}
\label{tab: mainresults}
\vspace{-5mm}
\end{table*}
\section{Experiments}
\label{sec4}

\subsection{Implementation Details}
\label{sec4b}
\textbf{Datasets:}
We conduct extensive experiments on the proposed \textbf{QAG-360K} and classic \textbf{CDVQA} datasets. For QAG-360K, the distribution of the original data is heavily biased, so we remove some questions from the dataset, while intentionally retaining the original real-world tendencies up to a tunable degree. The advantage of this scheme is to make the benchmark more challenging and less biased. After that, we randomly split the dataset into 70\% training, 10\% validation, and 20\% testing, making sure that all the questions about a given image appear in the same split. Besides, we use the CDVQA dataset to further validate the effectiveness of our method, which contains 2,968 image pairs, more than 120K question-answer pairs, and 6 land-cover categories.

\noindent \textbf{Evaluation Metrics:}
Besides the widely-used VQA metrics average accuracy (AA) and overall accuracy (OA), we further introduce mean IoU (mIoU) and overall IoU (oIoU) for CDQAG. 
Detailed definitions of metrics can be found in the Appendix C.1.

\noindent \textbf{Configurations:}
We employ the CLIP pre-trained Transformer \cite{radford2021learning} and ResNet-101 \cite{he2016deep} as the text and image encoders for all studies. The input image size and maximum sentence length are set to 512$\times$512 pixels and 40 tokens, respectively. Each multi-head attention layer and feed-forward network uses default settings \cite{dosovitskiy2020image}. For the loss function in \cref{eq:loss}, we set $\lambda_1$ = 0.2 and $\lambda_2$ = 1. In the main results (\cref{tab: mainresults}), the second column represents each type in \textit{\textbf{8C Questions}}: (CN: change or not; CtW: change to what; CfW: change from what; IN: increaxse or not; DN: decrease or not; LC: largest change; SC: smallest change; CR: change ratio). Following \cite{wang2022cris}, we train the model for 50 epochs using the AdamW \cite{loshchilov2017decoupled} optimizer with an initial learning rate of 1e-4. The learning rate is decreased by a factor of 0.1 at the 35th epoch. We train the model with a batch size of 64 on 4 NVIDIA RTX4090 with 24 GPU VRAM. During inference, we binarize the predicted results using a threshold of 0.35 to obtain the final outputs.

\noindent  \textbf{Baseline Methods:} 
Considering that existing methods rarely provide both textual and visual answers, we add the same text or mask prediction heads as ours to RES models (\eg, CRIS \cite{wang2022cris}, LAVT \cite{yang2022lavt}, CGFormer \cite{tang2023contrastive}, and ETRIS \cite{xu2023bridging}) and VQA models (\eg, RSVQA \cite{lobry2020rsvqa}, CDVQA \cite{yuan2022change}, and SOBA \cite{wang2024earthvqa}). Moreover, since the aforementioned methods are not specifically designed for change detection, to ensure a fair comparison, we concatenate the two images and feed them into the visual backbone in the same way for all methods (except that marked with symbol $\dagger$).

\subsection{Comparisons with State-of-the-art Methods}
\label{sec4d}
\noindent \textbf{Results on QAG-360K.}
In \cref{tab: mainresults}, we evaluate the performance of our proposed method in comparison with the state-of-the-art (SoTA) approaches on the QAG-360K test sets. Our method consistently outperforms competing methods in both the textual and visual answering tasks. Specifically, for classic textual answering, $\dagger$VisTA achieves the highest AA of \textbf{73.03\%} and OA of \textbf{77.35\%}. Most importantly, it provides absolute improvements of up to \textbf{6.76\%} on AA and \textbf{5.37\%} on OA over the cutting-edge VQA method SOBA \cite{wang2024earthvqa}, demonstrating its superior ability to effectively capture intricate change-related relationships and perform complex reasoning. In the intuitive visual answering, our model also attains the best performance, achieving an mIoU of \textbf{82.6\%} and an oIoU of \textbf{84.8\%}. Compared to CRIS \cite{wang2022cris}, CGFormer \cite{tang2023contrastive}, and ETRIS \cite{xu2023bridging}, which also use ResNet-101, our method achieves improvements of \textbf{14.71\%}/\textbf{13.30\%}/\textbf{9.57\%} on oIoU, respectively. $\dagger$VisTA reaches an impressive performance \textbf{78.39\%} on oIoU, reflecting a substantial advancement in comparison to the previous best method ETRIS (\textbf{13.25\%}/\textbf{11.73\%} for CtW/CfW). For the SC category, which requires finer spatial details, $\dagger$VisTA achieves a \textbf{6.90\%} improvement on oIoU over ETRIS. 
These performance gains highlight that our method can better model visual-language consistency, providing more precise and reliable visual evidence. Furthermore, compared to VisTA, the improved performance of $\dagger$VisTA shows the advantages of dual shared-weight encoders in effectively extracting and aligning change-related information across scenes.

\noindent \textbf{Results on CDVQA.}
We also compare our method with SoTA methods on the two CDVQA test subsets, as illustrated in \cref{tab:cdvqa}. Our method consistently outperforms the baselines in both AA and OA metrics. Specifically, on Test 1, our method achieves an OA of \textbf{73.1\%}, signifying a substantial improvement of \textbf{10.3\%} over RSVQA, \textbf{7.2\%} over CDVQA, and \textbf{3.9\%} over SOBA. On Test 2, our approach maintains its advantage with an AA of \textbf{66.0\%} and an OA of \textbf{69.5\%}, further demonstrating significant improvements over existing SoTA approaches. These results underscore the robustness and generalization capabilities of our model, particularly in challenging VQA scenarios. More experimental results are presented in Appendix C.3.

\begin{table}[!t]
\centering
\renewcommand{\arraystretch}{0.95}
\resizebox{\linewidth}{!}{
\footnotesize
    \begin{tabular}{c|c|cccc}
    \specialrule{.1em}{.05em}{.05em} 
    \multicolumn{2}{c|}{\multirow{2}{*}{Method}} & RSVQA \cite{lobry2020rsvqa}    & CDVQA \cite{yuan2022change} & SOBA \cite{wang2024earthvqa}     & \textbf{Ours} \\
    \multicolumn{2}{c|}{}                        & \textcolor{gray}{TGRS'21}      & \textcolor{gray}{TGRS'22}   & \textcolor{gray}{AAAI'24}        & - \\ 

    \specialrule{.1em}{.05em}{.05em} 
    \multirow{2}{*}{Test 1} 
    & AA & 52.1                           & 55.3                          & 60.3                          & \textbf{65.9} \\
    & OA & 62.8 \textcolor{blue}{(-10.3)} & 65.9 \textcolor{blue}{(-7.2)} & 69.2 \textcolor{blue}{(-3.9)} & \textbf{73.1} \\         
    \hline
    \multirow{2}{*}{Test 2} 
    & AA & 52.2                           & 55.4                          & 60.3                          & \textbf{65.9} \\
    & OA & 57.9 \textcolor{blue}{(-10.6)} & 61.1 \textcolor{blue}{(-7.4)} & 64.8 \textcolor{blue}{(-3.7)} & \textbf{68.5} \\                               
    \specialrule{.1em}{.05em}{.05em} 
    \end{tabular}}
\vspace{-3mm}
\caption{Comparison with the state-of-the-art methods on the two \textbf{CDVQA} test sets. The best performance are highlighted in \textbf{bold}.}
\label{tab:cdvqa}
\vspace{-5mm}
\end{table}

\begin{figure*}[!t]
	\centerline{\includegraphics[width=\linewidth]{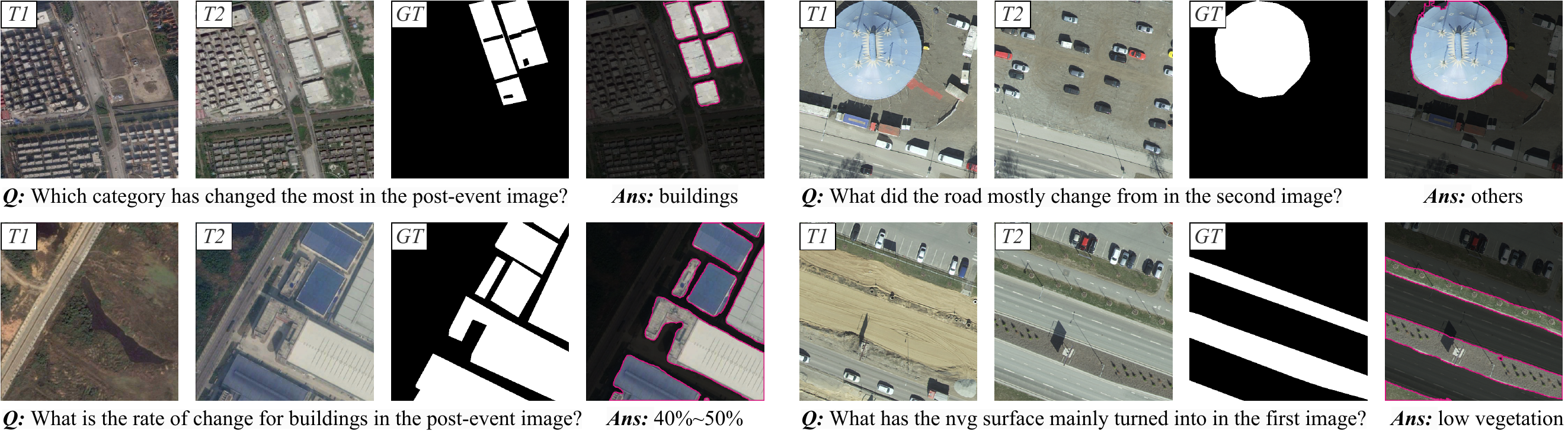}}
  \vspace{-3mm}
  \caption{Example results of our method on QAG-360K dataset.}
	\label{fig:visual-right}
  \vspace{-5mm}
\end{figure*}

\begin{figure}[!t]
	\centerline{\includegraphics[width=\linewidth]{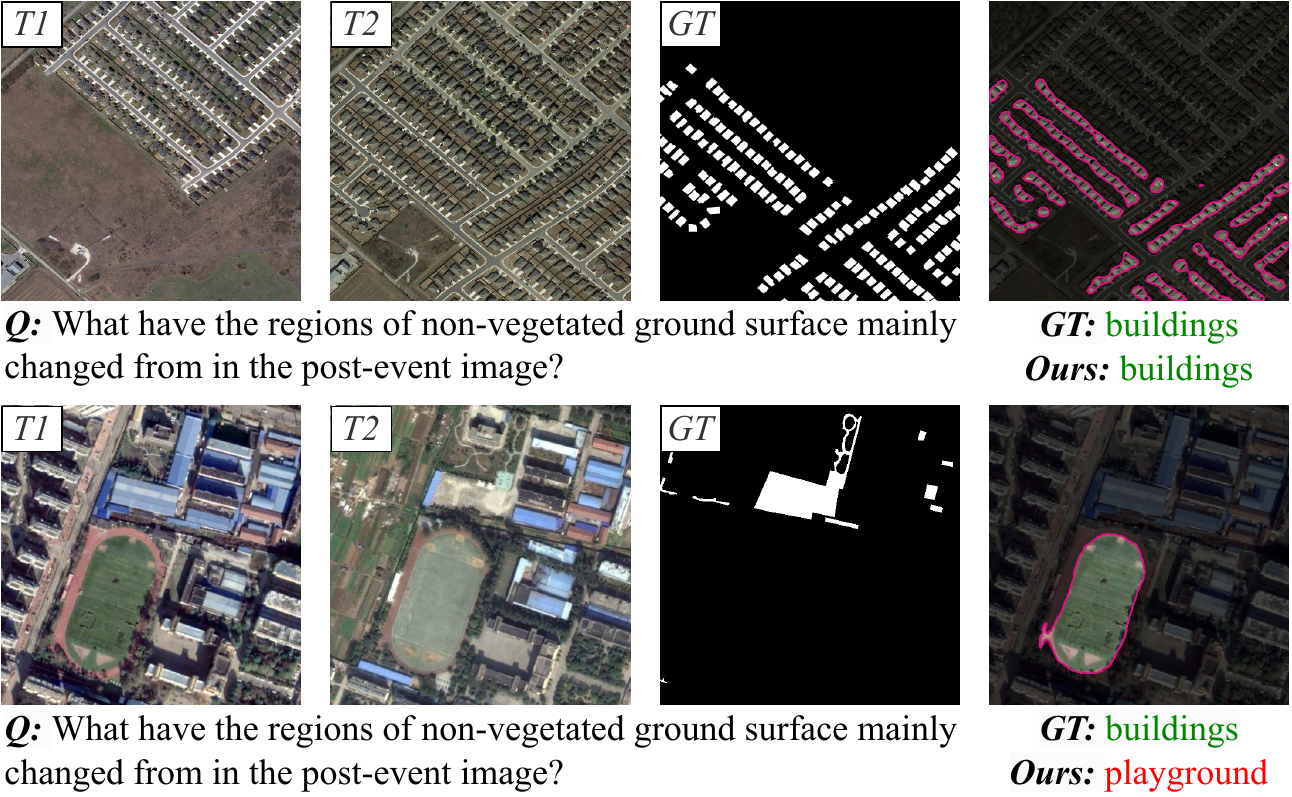}}
  \vspace{-3mm}
  \caption{Failure cases of our method on QAG-360K dataset.}
	\label{fig:visual-false}
  \vspace{-5mm}
\end{figure}

\noindent \textbf{Qualitative results}. 
As depicted in \cref{fig:visual-right}, our model accurately answers the change-related questions and provides intuitive and high-quality visual evidence. More visualizations are given in the Appendix C.4. We also show some failure cases in \cref{fig:visual-false}. In the first row, although the prediction is correct and the visual evidence aligns well with the textual answer, the regions of buildings are discrete units in the ground-truth (GT), while our model generates continuous masks with overly smooth boundaries. This discrepancy underscores the difficulty of accurately capturing the discrete nature of certain objects. In the second case, the color of the playground fades over time, leading to the model mistakenly classify it as a ``non-vegetated ground surface'' that closely resembles the visual change, resulting in an incorrect prediction. This further requires models to perceive subtle visual cues and comprehend context-sensitive differences for more accurate visual answers.

\subsection{Ablation Study}
\label{sec4e}

\noindent \textbf{Effect of VisTA's component.}
To validate the effectiveness of the proposed modules, we conduct ablation experiments on QAG-360K, which is summarized in \cref{tab:ablation_VisTA}.
In the baseline model, the features extracted from visual and text encoders are concatenated and then input to the segmentation and classification heads, respectively.
In Exp.2, language-guided feature aggregation (LGFA) is added onto the baseline model, which brings improvement of \textbf{1.61\%} OA and \textbf{3.04\%} oIoU.
Exp.3 and Exp.4 further investigate the importance of multi-stage reasoning in CDQAG task, demonstrating the effectiveness of fine-grained semantic information interaction.
Moreover, Exp.5 and Exp.6 highlight the benefit of auxiliary supervision via multimodal answer prediction, increasing the OA and oIoU of \textbf{4.90\%} and \textbf{3.97\%}.

\begin{figure}[!t]
	\centerline{\includegraphics[width=0.97\linewidth]{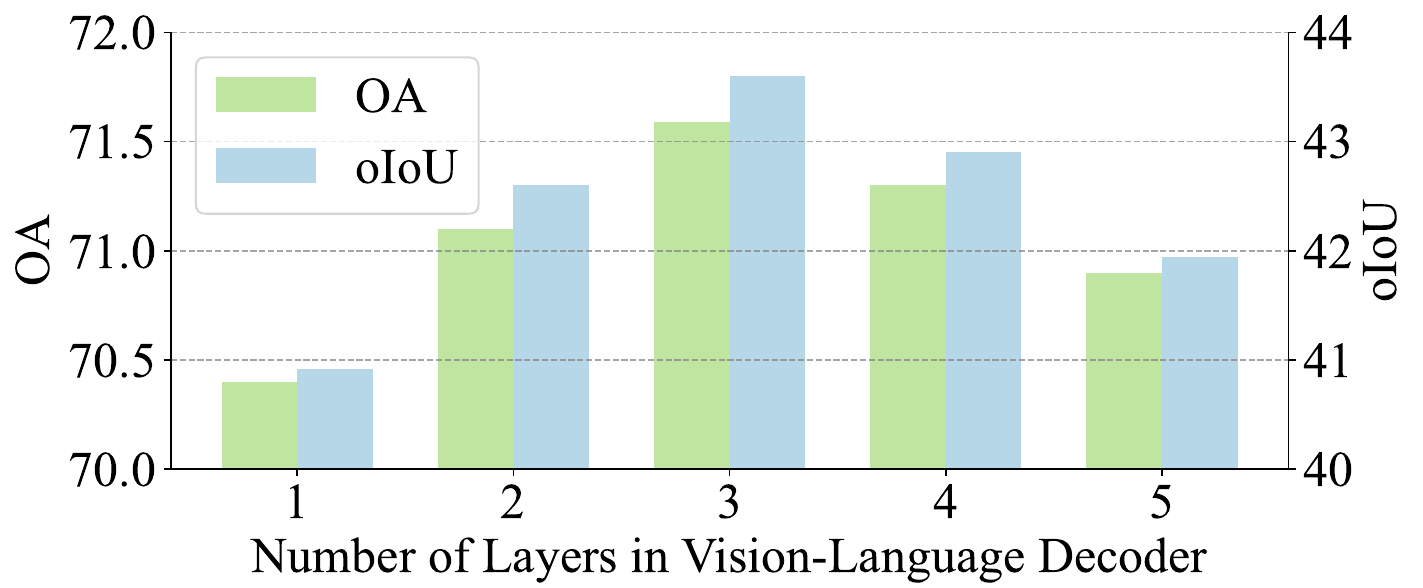}}
  \vspace{-3mm}
  \caption{Effectiveness of the number of V-L Decoder's layers.}
	\label{tab:ablation_numbers}
  \vspace{-5mm}
\end{figure}

\noindent \textbf{Numbers of Layers in Vision-Language Decoder.}
The impact of the number of layers in visual-language decoder is shown in \cref{tab:ablation_numbers}. When the visual representations are sequentially processed by more layers, the model consistently achieves higher OA and oIoU. However, the setting of n $\ge$ 4 introduces more parameters, which could increase the risk of over-fitting. Considering the performance and efficiency, we set n = 3 as the default in our framework.

\begin{table}[t]
\centering
\footnotesize
\renewcommand{\arraystretch}{0.95}
\begin{tabular}{c|l|ccccc}
    \specialrule{.1em}{.05em}{.05em} 
    Exp.         & Methods          & AA    & OA    & mIoU  & oIoU \\
    \hline 
    1 & Baseline                    & 62.58 & 68.29 & 28.73 & 37.51 \\
    2 & Exp.1 + LGFA                & 64.48 & 69.90 & 31.11 & 40.55 \\
    3 & Exp.2 + V-L Decoder         & 67.36 & 71.59 & 34.28 & 43.60 \\
    4 & Exp.3 + Q\&A Selector       & 69.77 & 73.88 & 37.90 & 46.02 \\
    \hline
    5 & w/o Textual Answering       & 58.21 & 70.86 & -     & -     \\
    6 & w/o Visual Answering        & -     & -     & 35.66 & 44.01 \\
    \hline
    7 & \textbf{VisTA} (Ours)   & \textbf{71.16} & \textbf{75.76} & \textbf{39.87} & \textbf{47.98}\\
    \specialrule{.1em}{.05em}{.05em} 
\end{tabular}
\vspace{-3mm}
\caption{Effectiveness of the proposed componets in VisTA.}
\label{tab:ablation_VisTA}
\vspace{-3mm}
\end{table}

\section{Conclusions and Limitations}
\label{sec5}
In this paper, we proposed the CDQAG benchmark, designed to provide intuitive visual explanations alongside textual answers. 
To facilitate research on CDQAG, we developed a large-scale dataset QAG-360K, and introduced a powerful baseline method VisTA, which sets new state-of-the-art performance on both the CDVQA and our QAG-360K datasets.
In addition, we present several analyses and insights for future research:
(1) Our experiments demonstrates that effective multimodal interaction is essential for analyzing diverse scenarios in change detection. 
This necessitates more efforts in achieving reliable complex reasoning.
(2) While our solution enhances question answering and grounding, delivering precise visual answers persists as a formidable challenge. Exploring solutions to this will greatly benefit the CDQAG task.
Finally, we highlight the importance of CDQAG for practical applications in remote sensing and hope it can inspire the community to drive progress in this domain.
{
    \small
    \bibliographystyle{ieeenat_fullname}
    \bibliography{main}

\begin{thebibliography}{45}
\providecommand{\natexlab}[1]{#1}
\providecommand{\url}[1]{\texttt{#1}}
\expandafter\ifx\csname urlstyle\endcsname\relax
  \providecommand{\doi}[1]{doi: #1}\else
  \providecommand{\doi}{doi: \begingroup \urlstyle{rm}\Url}\fi

\bibitem[Bernhard et~al.(2023)Bernhard, Strau{\ss}, and
  Schubert]{bernhard2023mapformer}
Maximilian Bernhard, Niklas Strau{\ss}, and Matthias Schubert.
\newblock Mapformer: Boosting change detection by using pre-change information.
\newblock In \emph{Proceedings of the IEEE/CVF International Conference on
  Computer Vision}, pages 16837--16846, 2023.

\bibitem[Chen and Shi(2020)]{chen2020spatial}
Hao Chen and Zhenwei Shi.
\newblock A spatial-temporal attention-based method and a new dataset for
  remote sensing image change detection.
\newblock \emph{Remote Sensing}, 12\penalty0 (10):\penalty0 1662, 2020.

\bibitem[Deng et~al.(2023)Deng, Yang, Liu, Chen, Zhou, Zhang, Li, and
  Ouyang]{deng2023transvg++}
Jiajun Deng, Zhengyuan Yang, Daqing Liu, Tianlang Chen, Wengang Zhou, Yanyong
  Zhang, Houqiang Li, and Wanli Ouyang.
\newblock Transvg++: End-to-end visual grounding with language conditioned
  vision transformer.
\newblock \emph{IEEE Transactions on Pattern Analysis and Machine
  Intelligence}, 2023.

\bibitem[Ding et~al.(2024)Ding, Zhang, Guo, Zhang, Liu, and
  Bruzzone]{ding2024joint}
Lei Ding, Jing Zhang, Haitao Guo, Kai Zhang, Bing Liu, and Lorenzo Bruzzone.
\newblock Joint spatio-temporal modeling for semantic change detection in
  remote sensing images.
\newblock \emph{IEEE Transactions on Geoscience and Remote Sensing}, 2024.

\bibitem[Dong et~al.(2024)Dong, Wang, Du, and Meng]{dong2024changeclip}
Sijun Dong, Libo Wang, Bo Du, and Xiaoliang Meng.
\newblock Changeclip: Remote sensing change detection with multimodal
  vision-language representation learning.
\newblock \emph{ISPRS Journal of Photogrammetry and Remote Sensing},
  208:\penalty0 53--69, 2024.

\bibitem[Dosovitskiy et~al.(2020)Dosovitskiy, Beyer, Kolesnikov, Weissenborn,
  Zhai, Unterthiner, Dehghani, Minderer, Heigold, Gelly,
  et~al.]{dosovitskiy2020image}
Alexey Dosovitskiy, Lucas Beyer, Alexander Kolesnikov, Dirk Weissenborn,
  Xiaohua Zhai, Thomas Unterthiner, Mostafa Dehghani, Matthias Minderer, Georg
  Heigold, Sylvain Gelly, et~al.
\newblock An image is worth 16x16 words: Transformers for image recognition at
  scale.
\newblock \emph{arXiv preprint arXiv:2010.11929}, 2020.

\bibitem[Gan et~al.(2017)Gan, Li, Li, Sun, and Gong]{gan2017vqs}
Chuang Gan, Yandong Li, Haoxiang Li, Chen Sun, and Boqing Gong.
\newblock Vqs: Linking segmentations to questions and answers for supervised
  attention in vqa and question-focused semantic segmentation.
\newblock In \emph{Proceedings of the IEEE/CVF International Conference on
  Computer Vision}, pages 1811--1820, 2017.

\bibitem[Goyal et~al.(2017)Goyal, Khot, Summers-Stay, Batra, and
  Parikh]{goyal2017making}
Yash Goyal, Tejas Khot, Douglas Summers-Stay, Dhruv Batra, and Devi Parikh.
\newblock Making the v in vqa matter: Elevating the role of image understanding
  in visual question answering.
\newblock In \emph{Proceedings of the IEEE/CVF Conference on Computer Vision
  and Pattern Recognition}, pages 6904--6913, 2017.

\bibitem[Gurari et~al.(2018)Gurari, Li, Stangl, Guo, Lin, Grauman, Luo, and
  Bigham]{gurari2018vizwiz}
Danna Gurari, Qing Li, Abigale~J Stangl, Anhong Guo, Chi Lin, Kristen Grauman,
  Jiebo Luo, and Jeffrey~P Bigham.
\newblock Vizwiz grand challenge: Answering visual questions from blind people.
\newblock In \emph{Proceedings of the IEEE/CVF Conference on Computer Vision
  and Pattern Recognition}, pages 3608--3617, 2018.

\bibitem[He et~al.(2016)He, Zhang, Ren, and Sun]{he2016deep}
Kaiming He, Xiangyu Zhang, Shaoqing Ren, and Jian Sun.
\newblock Deep residual learning for image recognition.
\newblock In \emph{Proceedings of the IEEE/CVF Conference on Computer Vision
  and Pattern Recognition}, pages 770--778, 2016.

\bibitem[Hu et~al.(2024)Hu, Li, Lu, Shao, He, Qiao, and Luo]{Hu_2024_CVPR}
Yutao Hu, Tianbin Li, Quanfeng Lu, Wenqi Shao, Junjun He, Yu Qiao, and Ping
  Luo.
\newblock Omnimedvqa: A new large-scale comprehensive evaluation benchmark for
  medical lvlm.
\newblock In \emph{Proceedings of the IEEE/CVF Conference on Computer Vision
  and Pattern Recognition}, pages 22170--22183, 2024.

\bibitem[Hudson and Manning(2019)]{hudson2019gqa}
Drew~A Hudson and Christopher~D Manning.
\newblock Gqa: A new dataset for real-world visual reasoning and compositional
  question answering.
\newblock In \emph{Proceedings of the IEEE/CVF Conference on Computer Vision
  and Pattern Recognition}, pages 6700--6709, 2019.

\bibitem[Johnson et~al.(2017)Johnson, Hariharan, Van Der~Maaten, Fei-Fei,
  Lawrence~Zitnick, and Girshick]{johnson2017clevr}
Justin Johnson, Bharath Hariharan, Laurens Van Der~Maaten, Li Fei-Fei, C
  Lawrence~Zitnick, and Ross Girshick.
\newblock Clevr: A diagnostic dataset for compositional language and elementary
  visual reasoning.
\newblock In \emph{Proceedings of the IEEE/CVF Conference on Computer Vision
  and Pattern Recognition}, pages 2901--2910, 2017.

\bibitem[Kirillov et~al.(2023)Kirillov, Mintun, Ravi, Mao, Rolland, Gustafson,
  Xiao, Whitehead, Berg, Lo, et~al.]{kirillov2023segment}
Alexander Kirillov, Eric Mintun, Nikhila Ravi, Hanzi Mao, Chloe Rolland, Laura
  Gustafson, Tete Xiao, Spencer Whitehead, Alexander~C Berg, Wan-Yen Lo, et~al.
\newblock Segment anything.
\newblock In \emph{Proceedings of the IEEE/CVF International Conference on
  Computer Vision}, pages 4015--4026, 2023.

\bibitem[Krishna et~al.(2017)Krishna, Zhu, Groth, Johnson, Hata, Kravitz, Chen,
  Kalantidis, Li, Shamma, et~al.]{krishna2017visual}
Ranjay Krishna, Yuke Zhu, Oliver Groth, Justin Johnson, Kenji Hata, Joshua
  Kravitz, Stephanie Chen, Yannis Kalantidis, Li-Jia Li, David~A Shamma, et~al.
\newblock Visual genome: Connecting language and vision using crowdsourced
  dense image annotations.
\newblock \emph{International Journal of Computer Vision}, 123:\penalty0
  32--73, 2017.

\bibitem[Lei et~al.(2019)Lei, Yu, Berg, and Bansal]{lei2019tvqa+}
Jie Lei, Licheng Yu, Tamara~L Berg, and Mohit Bansal.
\newblock Tvqa+: Spatio-temporal grounding for video question answering.
\newblock \emph{arXiv preprint arXiv:1904.11574}, 2019.

\bibitem[Li et~al.(2024)Li, Wang, Xu, Zhong, and Wang]{li2024language}
Ke Li, Di Wang, Haojie Xu, Haodi Zhong, and Cong Wang.
\newblock Language-guided progressive attention for visual grounding in remote
  sensing images.
\newblock \emph{IEEE Transactions on Geoscience and Remote Sensing}, 2024.

\bibitem[Lin et~al.(2017)Lin, Doll{\'a}r, Girshick, He, Hariharan, and
  Belongie]{lin2017feature}
Tsung-Yi Lin, Piotr Doll{\'a}r, Ross Girshick, Kaiming He, Bharath Hariharan,
  and Serge Belongie.
\newblock Feature pyramid networks for object detection.
\newblock In \emph{Proceedings of the IEEE/CVF Conference on Computer Vision
  and Pattern Recognition}, pages 2117--2125, 2017.

\bibitem[Liu et~al.(2024)Liu, Ma, Zhang, Wang, Ji, Sun, and Ji]{liu2024rotated}
Sihan Liu, Yiwei Ma, Xiaoqing Zhang, Haowei Wang, Jiayi Ji, Xiaoshuai Sun, and
  Rongrong Ji.
\newblock Rotated multi-scale interaction network for referring remote sensing
  image segmentation.
\newblock In \emph{Proceedings of the IEEE/CVF Conference on Computer Vision
  and Pattern Recognition}, pages 26658--26668, 2024.

\bibitem[Lobry et~al.(2020)Lobry, Marcos, Murray, and Tuia]{lobry2020rsvqa}
Sylvain Lobry, Diego Marcos, Jesse Murray, and Devis Tuia.
\newblock Rsvqa: Visual question answering for remote sensing data.
\newblock \emph{IEEE Transactions on Geoscience and Remote Sensing},
  58\penalty0 (12):\penalty0 8555--8566, 2020.

\bibitem[Loshchilov and Hutter(2017)]{loshchilov2017decoupled}
Ilya Loshchilov and Frank Hutter.
\newblock Decoupled weight decay regularization.
\newblock \emph{arXiv preprint arXiv:1711.05101}, 2017.

\bibitem[Ma et~al.(2024)Ma, Wang, Kong, Wang, Liu, Pei, and Zhao]{ma2024robust}
Jie Ma, Pinghui Wang, Dechen Kong, Zewei Wang, Jun Liu, Hongbin Pei, and
  Junzhou Zhao.
\newblock Robust visual question answering: Datasets, methods, and future
  challenges.
\newblock \emph{IEEE Transactions on Pattern Analysis and Machine
  Intelligence}, 2024.

\bibitem[Radford et~al.(2021)Radford, Kim, Hallacy, Ramesh, Goh, Agarwal,
  Sastry, Askell, Mishkin, Clark, et~al.]{radford2021learning}
Alec Radford, Jong~Wook Kim, Chris Hallacy, Aditya Ramesh, Gabriel Goh,
  Sandhini Agarwal, Girish Sastry, Amanda Askell, Pamela Mishkin, Jack Clark,
  et~al.
\newblock Learning transferable visual models from natural language
  supervision.
\newblock In \emph{International Conference on Machine Learning}, pages
  8748--8763. PMLR, 2021.

\bibitem[Ren et~al.(2015)Ren, Kiros, and Zemel]{ren2015exploring}
Mengye Ren, Ryan Kiros, and Richard Zemel.
\newblock Exploring models and data for image question answering.
\newblock \emph{Advances in Neural Information Processing Systems}, 28, 2015.

\bibitem[Ribeiro et~al.(2016)Ribeiro, Singh, and Guestrin]{ribeiro2016should}
Marco~Tulio Ribeiro, Sameer Singh, and Carlos Guestrin.
\newblock " why should i trust you?" explaining the predictions of any
  classifier.
\newblock In \emph{Proceedings of the 22nd ACM SIGKDD international conference
  on knowledge discovery and data mining}, pages 1135--1144, 2016.

\bibitem[Selvaraju et~al.(2017)Selvaraju, Cogswell, Das, Vedantam, Parikh, and
  Batra]{selvaraju2017grad}
Ramprasaath~R Selvaraju, Michael Cogswell, Abhishek Das, Ramakrishna Vedantam,
  Devi Parikh, and Dhruv Batra.
\newblock Grad-cam: Visual explanations from deep networks via gradient-based
  localization.
\newblock In \emph{Proceedings of the IEEE/CVF International Conference on
  Computer Vision}, pages 618--626, 2017.

\bibitem[Seo et~al.(2023)Seo, Lee, Jeon, and Seo]{seo2023self}
Minseok Seo, Hakjin Lee, Yongjin Jeon, and Junghoon Seo.
\newblock Self-pair: Synthesizing changes from single source for object change
  detection in remote sensing imagery.
\newblock In \emph{Proceedings of the IEEE/CVF Winter Conference on
  Applications of Computer Vision}, pages 6374--6383, 2023.

\bibitem[Tang et~al.(2023)Tang, Zheng, Shi, and Yang]{tang2023contrastive}
Jiajin Tang, Ge Zheng, Cheng Shi, and Sibei Yang.
\newblock Contrastive grouping with transformer for referring image
  segmentation.
\newblock In \emph{Proceedings of the IEEE/CVF Conference on Computer Vision
  and Pattern Recognition}, pages 23570--23580, 2023.

\bibitem[Tian et~al.(2020)Tian, Ma, Zheng, and Zhong]{tian2020hi}
Shiqi Tian, Ailong Ma, Zhuo Zheng, and Yanfei Zhong.
\newblock Hi-ucd: A large-scale dataset for urban semantic change detection in
  remote sensing imagery.
\newblock \emph{arXiv preprint arXiv:2011.03247}, 2020.

\bibitem[Toker et~al.(2022)Toker, Kondmann, Weber, Eisenberger, Camero, Hu,
  Hoderlein, {\c{S}}enaras, Davis, Cremers, et~al.]{toker2022dynamicearthnet}
Aysim Toker, Lukas Kondmann, Mark Weber, Marvin Eisenberger, Andr{\'e}s Camero,
  Jingliang Hu, Ariadna~Pregel Hoderlein, {\c{C}}a{\u{g}}lar {\c{S}}enaras,
  Timothy Davis, Daniel Cremers, et~al.
\newblock Dynamicearthnet: Daily multi-spectral satellite dataset for semantic
  change segmentation.
\newblock In \emph{Proceedings of the IEEE/CVF Conference on Computer Vision
  and Pattern Recognition}, pages 21158--21167, 2022.

\bibitem[Urooj et~al.(2021)Urooj, Kuehne, Duarte, Gan, Lobo, and
  Shah]{urooj2021found}
Aisha Urooj, Hilde Kuehne, Kevin Duarte, Chuang Gan, Niels Lobo, and Mubarak
  Shah.
\newblock Found a reason for me? weakly-supervised grounded visual question
  answering using capsules.
\newblock In \emph{Proceedings of the IEEE/CVF Conference on Computer Vision
  and Pattern Recognition}, pages 8465--8474, 2021.

\bibitem[Wang et~al.(2024{\natexlab{a}})Wang, Dong, Li, and
  Chen]{wang2024kernel}
Di Wang, Fuyu Dong, Ke Li, and Duo Chen.
\newblock Kernel-adaptive change detection network in remote sensing imagery.
\newblock In \emph{IEEE International Geoscience and Remote Sensing Symposium},
  pages 10192--10196. IEEE, 2024{\natexlab{a}}.

\bibitem[Wang et~al.(2024{\natexlab{b}})Wang, Zheng, Chen, Ma, and
  Zhong]{wang2024earthvqa}
Junjue Wang, Zhuo Zheng, Zihang Chen, Ailong Ma, and Yanfei Zhong.
\newblock Earthvqa: Towards queryable earth via relational reasoning-based
  remote sensing visual question answering.
\newblock In \emph{Proceedings of the AAAI Conference on Artificial
  Intelligence}, pages 5481--5489, 2024{\natexlab{b}}.

\bibitem[Wang et~al.(2022)Wang, Lu, Li, Tao, Guo, Gong, and Liu]{wang2022cris}
Zhaoqing Wang, Yu Lu, Qiang Li, Xunqiang Tao, Yandong Guo, Mingming Gong, and
  Tongliang Liu.
\newblock Cris: Clip-driven referring image segmentation.
\newblock In \emph{Proceedings of the IEEE/CVF Conference on Computer Vision
  and Pattern Recognition}, pages 11686--11695, 2022.

\bibitem[Wu et~al.(2023)Wu, Du, and Zhang]{wu2023fully}
Chen Wu, Bo Du, and Liangpei Zhang.
\newblock Fully convolutional change detection framework with generative
  adversarial network for unsupervised, weakly supervised and regional
  supervised change detection.
\newblock \emph{IEEE Transactions on Pattern Analysis and Machine
  Intelligence}, 45\penalty0 (8):\penalty0 9774--9788, 2023.

\bibitem[Xu et~al.(2023)Xu, Chen, Zhang, Song, Wan, and Li]{xu2023bridging}
Zunnan Xu, Zhihong Chen, Yong Zhang, Yibing Song, Xiang Wan, and Guanbin Li.
\newblock Bridging vision and language encoders: Parameter-efficient tuning for
  referring image segmentation.
\newblock In \emph{Proceedings of the IEEE/CVF International Conference on
  Computer Vision}, pages 17503--17512, 2023.

\bibitem[Yang et~al.(2021)Yang, Xia, Liu, Du, Yang, Pelillo, and
  Zhang]{yang2021asymmetric}
Kunping Yang, Gui-Song Xia, Zicheng Liu, Bo Du, Wen Yang, Marcello Pelillo, and
  Liangpei Zhang.
\newblock Asymmetric siamese networks for semantic change detection in aerial
  images.
\newblock \emph{IEEE Transactions on Geoscience and Remote Sensing},
  60:\penalty0 1--18, 2021.

\bibitem[Yang et~al.(2022{\natexlab{a}})Yang, Xu, Yuan, Liu, Li, and
  Hu]{yang2022improving}
Li Yang, Yan Xu, Chunfeng Yuan, Wei Liu, Bing Li, and Weiming Hu.
\newblock Improving visual grounding with visual-linguistic verification and
  iterative reasoning.
\newblock In \emph{Proceedings of the IEEE/CVF Conference on Computer Vision
  and Pattern Recognition}, pages 9499--9508, 2022{\natexlab{a}}.

\bibitem[Yang et~al.(2022{\natexlab{b}})Yang, Wang, Tang, Chen, Zhao, and
  Torr]{yang2022lavt}
Zhao Yang, Jiaqi Wang, Yansong Tang, Kai Chen, Hengshuang Zhao, and Philip~HS
  Torr.
\newblock Lavt: Language-aware vision transformer for referring image
  segmentation.
\newblock In \emph{Proceedings of the IEEE/CVF Conference on Computer Vision
  and Pattern Recognition}, pages 18155--18165, 2022{\natexlab{b}}.

\bibitem[Ye et~al.(2022)Ye, Tian, Yan, Yang, Wang, Zhang, He, and
  Lin]{ye2022shifting}
Jiabo Ye, Junfeng Tian, Ming Yan, Xiaoshan Yang, Xuwu Wang, Ji Zhang, Liang He,
  and Xin Lin.
\newblock Shifting more attention to visual backbone: Query-modulated
  refinement networks for end-to-end visual grounding.
\newblock In \emph{Proceedings of the IEEE/CVF Conference on Computer Vision
  and Pattern Recognition}, pages 15502--15512, 2022.

\bibitem[Yuan et~al.(2022)Yuan, Mou, Xiong, and Zhu]{yuan2022change}
Zhenghang Yuan, Lichao Mou, Zhitong Xiong, and Xiao~Xiang Zhu.
\newblock Change detection meets visual question answering.
\newblock \emph{IEEE Transactions on Geoscience and Remote Sensing},
  60:\penalty0 1--13, 2022.

\bibitem[Zheng et~al.(2021)Zheng, Ma, Zhang, and Zhong]{zheng2021change}
Zhuo Zheng, Ailong Ma, Liangpei Zhang, and Yanfei Zhong.
\newblock Change is everywhere: Single-temporal supervised object change
  detection in remote sensing imagery.
\newblock In \emph{Proceedings of the IEEE/CVF International Conference on
  Computer Vision}, pages 15193--15202, 2021.

\bibitem[Zheng et~al.(2022)Zheng, Zhong, Tian, Ma, and
  Zhang]{zheng2022changemask}
Zhuo Zheng, Yanfei Zhong, Shiqi Tian, Ailong Ma, and Liangpei Zhang.
\newblock Changemask: Deep multi-task encoder-transformer-decoder architecture
  for semantic change detection.
\newblock \emph{ISPRS Journal of Photogrammetry and Remote Sensing},
  183:\penalty0 228--239, 2022.

\bibitem[Zheng et~al.(2023)Zheng, Tian, Ma, Zhang, and
  Zhong]{zheng2023scalable}
Zhuo Zheng, Shiqi Tian, Ailong Ma, Liangpei Zhang, and Yanfei Zhong.
\newblock Scalable multi-temporal remote sensing change data generation via
  simulating stochastic change process.
\newblock In \emph{Proceedings of the IEEE/CVF International Conference on
  Computer Vision}, pages 21818--21827, 2023.

\bibitem[Zhu et~al.(2016)Zhu, Groth, Bernstein, and Fei-Fei]{zhu2016visual7w}
Yuke Zhu, Oliver Groth, Michael Bernstein, and Li Fei-Fei.
\newblock Visual7w: Grounded question answering in images.
\newblock In \emph{Proceedings of the IEEE/CVF Conference on Computer Vision
  and Pattern Recognition}, pages 4995--5004, 2016.

\end{thebibliography}
}

\end{document}